\documentclass[runningheads]{llncs}
\usepackage{graphicx}
\usepackage{comment}
\usepackage{amsmath,amssymb} 
\usepackage{color}

\usepackage{epsfig}
\usepackage{bbm}
\usepackage{commath}
\usepackage{url}
\usepackage{enumitem}
\usepackage[linesnumbered,boxed,ruled,commentsnumbered]{algorithm2e}
\usepackage{caption}
\usepackage{multirow}
\usepackage{lipsum}

\usepackage{bbm}
\usepackage{enumitem}

\makeatletter
\renewcommand\section{\@startsection{section}{1}{\z@}%
                       {-4\p@ \@plus -4\p@ \@minus -4\p@}%
                       {6\p@ \@plus 4\p@ \@minus 4\p@}%
                       {\normalfont\large\bfseries\boldmath
                        \rightskip=\z@ \@plus 8em\pretolerance=10000 }}
\renewcommand\subsection{\@startsection{subsection}{2}{\z@}%
                       {-4\p@ \@plus -4\p@ \@minus -4\p@}%
                       {6\p@ \@plus 4\p@ \@minus 4\p@}%
                       {\normalfont\normalsize\bfseries\boldmath
                        \rightskip=\z@ \@plus 8em\pretolerance=10000 }}
\renewcommand\subsubsection{\@startsection{subsubsection}{3}{\z@}%
                       {-2\p@ \@plus -4\p@ \@minus -4\p@}%
                       {-1.5em \@plus -0.22em \@minus -0.1em}%
                       {\normalfont\normalsize\bfseries\boldmath}}
\makeatother
\begin{document}
\pagestyle{headings}
\mainmatter
\def\ECCVSubNumber{3135}  

\title{High-Resolution Image Inpainting with Iterative Confidence Feedback and Guided Upsampling} 

\titlerunning{Inpainting with Iterative Confidence Feedback and Guided Upsampling}
%
%
\author{Yu Zeng\inst{1} \and
Zhe Lin\inst{2} \and
Jimei Yang\inst{2} \and
Jianming Zhang\inst{2} \and
Eli Shechtman\inst{2} \and
Huchuan Lu\inst{1}
}
\authorrunning{Y. Zeng et al.}
%
\institute{Dalian University of Technology, China\\
\email{zengxianyu18@qq.com,lhchuan@dlut.edu.cn}\\
\and
Adobe Research, USA\\
\email{\{zlin,jimyang,jianmzha,elishe\}@adobe.com}}

\maketitle
\begin{abstract}
Existing image inpainting methods often produce artifacts when dealing with large holes in real applications. To address this challenge, we propose an iterative inpainting method with a feedback mechanism. Specifically, we introduce a deep generative model which not only outputs an inpainting result but also a corresponding confidence map. Using this map as feedback, it progressively fills the hole by trusting only high-confidence pixels inside the hole at each iteration and focuses on the remaining pixels in the next iteration. As it reuses partial predictions from the previous iterations as known pixels, this process gradually improves the result. In addition, we propose a guided upsampling network to enable generation of high-resolution inpainting results. We achieve this by extending the Contextual Attention module~[39] to borrow high-resolution feature patches in the input image.
Furthermore, to mimic real object removal scenarios, we collect a large object mask dataset and synthesize more realistic training data that better simulates user inputs. Experiments show that our method significantly outperforms existing methods in both quantitative and qualitative evaluations. More results and Web APP are available at https://zengxianyu.github.io/iic.
%
\end{abstract}
\section{Introduction}
Image inpainting is a task of reconstructing missing regions in an image. It is an important problem in computer vision and an essential functionality in many imaging and graphics applications, \textit{e.g. } object removal, image restoration, manipulation, re-targeting, compositing, and image-based rendering~\cite{barnes2009patchmatch,levin2004seamless,park2017transformation} 

Classical inpainting methods such as~\cite{efros2001image,kwatra2005texture,barnes2009patchmatch} typically rely on the principle of borrowing example patches from known regions or external database images and pasting them into the holes. These methods are quite effective for easy cases with small holes or uniform textured background. They can also handle high-resolution images efficiently. However, due to the lack of high-level structural understanding and ability to generate novel contents, they often fail to produce realistic results when the hole is large. 

\begin{figure}[t]
\begin{center}
\includegraphics[width=.9\linewidth]{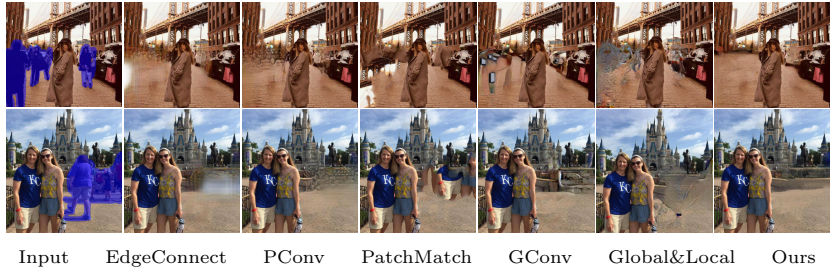}\\
\makebox[\linewidth][s]{\scriptsize{ { } { } Input { } EdgeConnect { } PConv { } PatchMatch { } GConv { } Global\&Local { } Ours { } { } }}
\end{center}
  \captionof{figure}{\footnotesize Results of state-of-the-art methods on real object removal requests~\cite{remove_people_on_side,remove_woman_in_purple}}.
\label{fig:teaser}
\end{figure}

Deep learning has achieved great success on various dense prediction problems~\cite{ding2018context_cvpr,ding2019semantic,ding2019boundary,ding2020phraseclick,zeng2020high,zeng2018learning,zeng2019multi,zeng2019joint,wang2020sdc}. Recent research effort on inpainting has shifted towards a data-driven learning-based approach~\cite{pathak2016context,iizuka2017globally,yu2018generative,liu2018image,yu2019free}. These methods train a deep neural network to directly predict the inpainting result given a corrupted image and hole channel as input. The original images before corruption are used as the ground truth to train the network. To generate visually realistic results with sufficient texture details, they often use an adversarial loss based on GANs~\cite{goodfellow2014generative} in addition to a reconstruction loss. These deep generative models show significant improvements on filling holes in complex images but often produce visual artifacts, especially when the hole is large. For large holes, the reconstruction loss is less effective due to the increased ambiguity, leading to undesired predictions during testing 
as shown in Fig.~\ref{fig:teaser}.

In this paper, we aim to address the challenge of filling large holes in high resolution images for real image editing applications, \textit{e.g.}, object removal. We observe that in the failure cases of existing approaches, despite the artifacts, there often exist sub-regions with good predictions. If we trust the good part and treat the remaining region as a new hole and run the model again, then the hole become progressively smaller and the model can produce better results. Inspired by this observation, we propose a novel iterative inpainting method with a feedback mechanism. Our method is based on a deep generative model which not only outputs an inpainting result but also a corresponding confidence map. 

The model is encouraged to generate a confidence map that highlights pixels where the prediction error is likely small and can help overcome the prediction ambiguity. Using this confidence map as feedback, our model is trained to progressively fill the hole by trusting only high-confidence pixels inside the hole at each iteration and update the remaining pixels in the next iteration. By predicting what portion of the hole was filled successfully in the previous iteration and using those pixels as ``known'', our model can gradually improve the result when filling large holes. 
The proposed confidence prediction scheme is general and can be potentially attached to any deep generative inpainting model. 


To generate high-quality inpainting results at high-resolution, we propose a guided inpainting upsampling network as a post-processing method. We achieve this by extending the Contextual Attention module~\cite{yu2018generative} to borrow known high-resolution feature patches in the input image based on the patch similarities computed on the result for down-sampled input. 
The motivation is that it is easier to train a deep network to generate globally coherent structures for down-sampled inputs as effective receptive fields of neurons are larger; while the surrounding regions of the high-resolution input can be used to enhance fine-grained texture details inside the hole.
In this way, our method decouples high-level structural understanding and low-level texture reconstruction, and can produce results that are both semantically plausible and visually realistic at high resolution. 

On the data side, previous methods construct training data by synthesizing square~\cite{pathak2016context,iizuka2017globally,yu2018generative} or irregular holes~\cite{liu2018image,yu2019free}.
However, in real applications for removing undesirable objects or scene segments, the hole-filling regions are more likely to be object-shaped. 
To mimic real use cases, we collect a large set of images with object-shaped holes. We synthesize our data to include both of the following two common situations: 1) holes overlap the foreground objects to simulate the distracting objects occluding the foreground, and 2) holes appear only in the background to simulate the case of unwanted objects occluded by salient foreground objects. 

In summary, our contributions are three-fold:
\begin{itemize}[noitemsep]
\item We address the challenge of completing large missing regions in images by proposing an iterative inpainting method with a confidence feedback loop. 
\item We propose a guided up-sampling network as a post processing step to enable generation of high-resolution inpainting results. 
\item We introduce a new procedure to synthesize training data for building deep generative models for real object removal applications.
\end{itemize}

\section{Related work}
Earlier image inpainting methods rely on existing content to fill the holes. Diffusion-based methods~\cite{ballester2001filling,bertalmio2000image} propagate neighboring appearances to the target holes, but they often generate significant artifacts when the holes are large or texture variation is severe. 
Patch-based methods~\cite{efros2001image,kwatra2005texture,barnes2009patchmatch} search for most similar patches from valid regions to complete missing regions. 
Drori~\textit{et al.}~\cite{drori2003fragment} propose to iteratively fill missing regions from high to low confidence with similar patches. Although they also use a map to determine the region to fill in each iteration, the map is predefined based on spatial distances from unknown pixels to their closest valid pixels. 
The above methods use real image patches sampled from the input to fill the holes and can often generate high-quality results. However, they lack high-level structural understanding and cannot generate entirely new content that does not exist in the input image. Thus, their results may not be semantically consistent to regions surrounding the holes. 

By learning from a large corpus of data, deep learning based inpainting methods can understand the semantic structure of the input image and hence can handle more difficult cases. To produce sharper results, these methods typically adopt adversarial training inspired by GANs~\cite{goodfellow2014generative}. Pathak~\textit{et al.}~\cite{pathak2016context} made a first attempt to use a convolutional neural network (CNN) for hole filling. 
Iizuka~\textit{et al.}~\cite{iizuka2017globally} use two discriminators for adversarial training to make the inpainted content both locally and globally consistent.  Yu~\textit{et al.}~\cite{yu2018generative} propose a deep generative model with contextual attention to explicitly utilize surrounding image features as references in the latent feature space. Zeng~\textit{et al.}~\cite{zeng2019learning} propose to use region affinity from a high-level feature map to guide the completion of missing regions in the previous low-level feature map of a single input. Our upsampling network is similar in spirit of using coarse scale information to guide the generation of fine-grained details but different in architecture and functionality; it upsamples a low-resoluiotn results by filling the fine-grained details from the high-resolution input. Yang~\textit{et al.}~\cite{yang2017neuralpatch} upsamples the results of a similar network with a neural patch search and vote approach followed by an optimization. Our method uses a related neural patch-vote approach but avoids the slow optimization.
The above methods use square holes in their training data, which causes a bias to rectangular holes. To address this, Liu~\textit{et al.}~\cite{liu2018image} collect estimated occlusion/dis-occlusion masks between two consecutive frames of videos and use them to generate holes for training. The resulting masks are highly irregular and do not represent well holes typical to an image inpainting task.
They also propose partial convolution layers to infer missing pixels conditioned only on valid pixels. 
Yu~\textit{et al.}~\cite{yu2019free} introduce free-form masks by simulating random strokes and generalizes partial convolution to gated convolution that learns to select features for each channel at each spatial location across all layers. Although these irregular holes lead to more diverse samples, they do not represent well real inpainting use-cases. 

Most recently, a few works have been introduced to study progressive inpainting models. Zhang~\textit{et al.}~\cite{zhang2018semantic} adopt a UNet generator with an LSTM in the bottleneck. It takes a sequence of inputs with large to small holes in the image center and generates a sequence of corresponding outputs. Guo~\textit{et al.}~\cite{guo2019progressive} propose to gradually fill a hole using consecutive residual blocks. They use partial convolutions in these blocks and update the hole mask according to the invalid pixels selected by partial convolutions. Oh~\textit{et al.}~\cite{oh2019onion} propose an onion-peel network that progressively fills the hole from the hole boundary for video hole filling. All of the above methods fill the holes from the boundary to inner regions in a {\em predefined} sequence. Different from them, our proposed method jointly predicts a confidence map when generating an inpainting result. Using the confidence map from the previous iteration as feedback, it can automatically detect regions with bad fill to revise in following iterations. To our best knowledge, it is the first attempt to model confidence of predictions in inpainting and the first iterative inpainting method to fill holes with a confidence-driven feedback loop.

\section{Approach}
Our inpainting method consists of two models: an iterative inpainting model (Fig.~\ref{fig:overall_split1}~(a)) with confidence feedback and a guided upsampling network ((Fig.~\ref{fig:overall_split1}~(b)) that upsamples a low-resolution result by factor of $2$ using the high resolution (HR) input as guidance. 
In this section, we first describe how we prepare data for building and evaluating our model, and then introduce the details of our iterative inpainting model and guided upsampling network. 

\subsection{Data generation}
\label{sec_data}
\begin{figure}[t]
\begin{center}
\includegraphics[width=.9\linewidth]{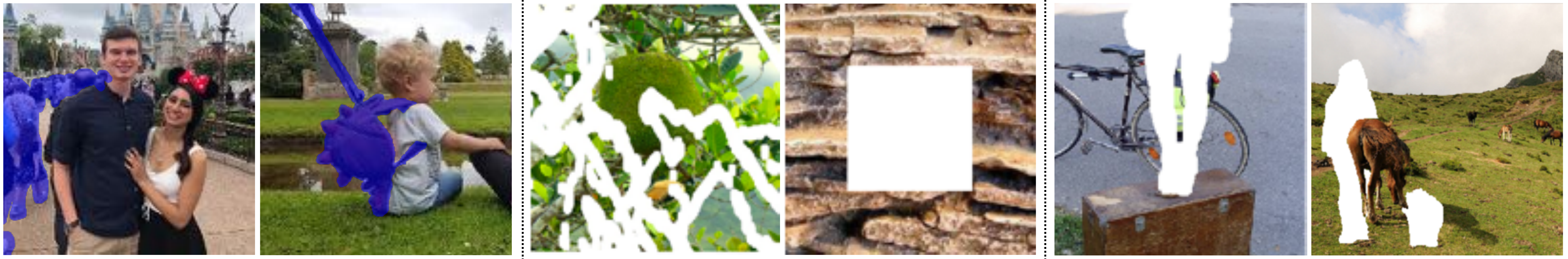}\\
\makebox[\linewidth][s]{\scriptsize{ { } Real{\color{white}w}request { } Previous{\color{white}w}approaches { } Ours { } }}
\end{center}
   \caption{\small Comparison of input with holes. The first two columns are real object removal requests on the Web~\cite{man_woman_bald_head,child_backpack_lead}. 
   The second two are from PConv~\cite{liu2018image} and ContextAttention~\cite{yu2018generative}, respectively. The right two are our samples with object-shaped holes. }
\label{fig:samples}
\end{figure}

Previous approaches to image inpainting typically construct their training and testing data pairs by corrupting the original images with square-shaped~\cite{yu2018generative,pathak2016context,iizuka2017globally} or highly irregular holes~\cite{liu2018image,yu2019free}, as shown in the first two columns in Fig.~\ref{fig:samples}. Images with holes are the input and the original images are the corresponding ground-truth. However, in real-world inpainting use cases such as distracting region removal, the regions users typically want to remove are objects or scene segments, which are rarely of square or highly irregular shapes, as shown in the middle two columns of Fig.~\ref{fig:samples}.

To mimic a more typical image editing scenario, in this work, we synthesize training samples with realistic holes. We collect 82,020 object masks from densely annotated segmentation datasets, including video segmentation~\cite{Caelles_arXiv_2019}, semantic segmentation~\cite{everingham2010pascal}, salient object segmentation~\cite{LiYu16,WangDRFI2017,fan2018SOC}, and human parsing~\cite{ATR}. The object masks from semantic segmentation datasets are from various classes 
and have different shapes and sizes. Salient object segmentation datasets contain large objects that form samples with large holes in our training data. We also use human parsing datasets to generate human-shaped masks as removing distracting people from photos is a common task. 

Finally, we use a mix of random strokes~\cite{yu2019free} and the object masks as holes to create a more diverse training dataset and overcome a bias towards object shaped holes. The images for synthesizing training samples are from two sources: the Places2~\cite{zhou2017places} dataset and salient object segmentation dataset~\cite{xiong2019foreground}. More specifically, we collect 61,525 images with pixel-level annotations of salient objects. We use 1,000 of them as testing samples, and the rest (60,525) are merged with Places2 for training and validation. For images in Places2, we sample randomly the location of the holes so they can appear in any region and may overlap with the main object. For images originating from salient object segmentation datasets, we subtract from the holes the intersection area with the salient objects. This is to simulate the case of removing distracting regions occluded by salient objects. As shown in the last two columns of Fig.~\ref{fig:samples}, such generated samples with holes are more similar to real cases than those of previous approaches. 

\subsection{Inpainting model}
\begin{figure}[t]
\begin{center}
  \includegraphics[width=0.5\linewidth]{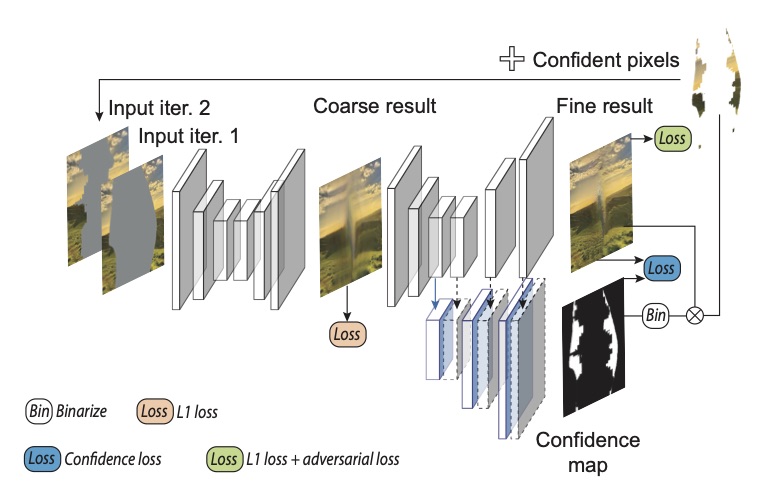}
  \includegraphics[width=0.45\linewidth]{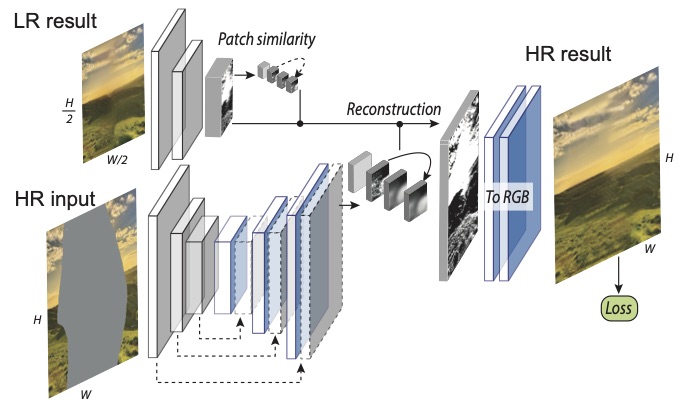}
  \makebox[\linewidth][s]{\scriptsize{{ } (a) { } (b) { }}}
\end{center}
  \caption{\small The overall structure. (a) Iterative inpainting with confidence feedback. (b) Guided upsampling.}
\label{fig:overall_split1}
\end{figure}
We adopt a generative approach based on generative adversarial networks (GANs)~\cite{goodfellow2014generative}. Thus our model has a generator and a discriminator. Fig.~\ref{fig:overall_split1}~(a) illustrates an overall structure of the generator. It is a cascade of a coarse and a fine networks, similar to~\cite{yu2018generative}. The coarse network takes an incomplete image and the corresponding hole mask as input to produce a coarse completed image. Then, the coarse result is passed to the fine network to generate a final completed image and a confidence map. 
The fine network has one encoder and two decoders: an image decoder that predicts the inpainting image result, and a confidence decoder that returns a corresponding confidence map of the predicted image. To make confidence prediction aware of the full generation process, we let the confidence decoder take as input all the feature layers up to the bottleneck of the image decoder, as illustrated by the dashed blocks in Fig.~\ref{fig:overall_split1}~(a). 

We use a PatchGAN discriminator~\cite{isola2017image} with spectral normalization as in~\cite{yu2019free} for adversarial training. It takes as input either the inpainted image or the ground-truth image and classifies each patch of the input image as real or fake. Its output is a score map rather than a single score, where each element corresponds to a local region of the input image covered by its receptive field. 

\subsubsection{Generative inpainting loss}
We train our model on the realistic hole dataset described in Sec.~\ref{sec_data}. We use L1 reconstruction loss on the coarse level. On the fine level, we use both L1 and a hinge adversarial loss with spectral normalization~\cite{miyato2018spectral} applied on the discriminator. The loss for the discriminator $D$ is: 
\begin{equation}
\mathcal{L}_D = \mathbb{E}_{x\sim p_{data}(x)}[\mbox{ReLU}(\mathbbm{1}-D(x))] + \mathbb{E}_{z\sim p_{z}(z)}[\mbox{ReLU}(\mathbbm{1}+D(G(z)\circ m+z))],  \\
\end{equation}
where $x$ denotes the real (ground truth) image and $z$ represents the incomplete image of which the pixels inside the hole are set to zero; $m$ represents the hole mask, in which the pixels having value one belong to the hole; $G(\cdot)$ represents the image decoder; $\circ$ denotes element-wise multiplication; the inpainting result $G(z)\circ m+z$ is composed by the generated content $G(z)$ inside the hole and the original content $z$ outside the hole. Let $y$ denote the output of the image decoder, \textit{i.e.} $y=G(z)$, then the loss for the inpainting result is:
\begin{equation}
\label{eq_image_loss}
\mathcal{L}_{G} = \mathbb{E}_{z,x\sim p(z,x)}[L(y)], \mbox{where} L(y) = \mbox{ReLU}(\mathbbm{1}-D(y\circ m+z)) + \norm{y-x}_1.
\end{equation}

\subsubsection{Confidence prediction loss}
We make the confidence decoder detect good regions by using its output map as spatial attention on the predicted image when calculating the loss. Let $c$ denote the confidence map, i.e., output of the confidence decoder of which each element is constrained to $[0, 1]$ by a sigmoid function, we define the following loss for the confidence decoder:
\begin{equation}
\label{eq_mask_loss}
\mathcal{L}_{C} = \mathbb{E}_{z,x\sim p(z,x)}[L(y\circ c+x\circ(1-c)) + \lambda (\norm{(\mathbbm{1}-c)\circ m}_1+\norm{(\mathbbm{1}-c)\circ m}_2)],  
\end{equation}
where $\lambda$ is a hyperparameter controlling the size of the confident area. We set it to 0.1 in all evaluations but also provide sensitivity analysis of it in the experiments. 

\begin{figure}[t]
\begin{center}
  \includegraphics[width=0.5\linewidth]{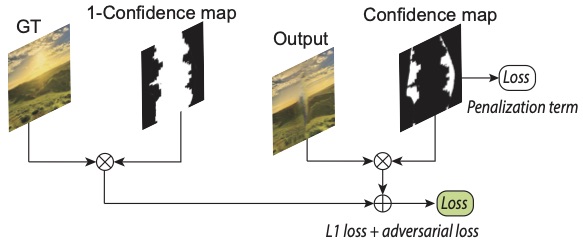}
\end{center}
  \caption{\small Loss for confidence prediction. }
\label{fig:loss_conf}
\end{figure}

To minimize $\mathcal{L}_{C}$, the map $c$ should highlight confidence regions, i.e. pixels contributing less to the overall loss. To prove this, we assume that: (1) $L(y)$ can be written as the summation over local pixel-wise losses, \textit{i.e.}  
$L(y) = \sum_{i\in\mathcal{H}}l(y_i)$ where $l(y_i)\ge 0$ is an unknown local loss function, $\mathcal{H}$ is the index set of pixels inside the holes and $y_i$ is a pixel of the generator output $y$, and (2) $l(x_i)=0$ for every ground-truth pixel $x_i$. For simplification, we consider $c$ as binary and let $\mathcal{C}$ to be the index set of non-zero elements of $c$, then $\mathcal{L}_{C}$ can be re-written as:
\begin{equation}
\label{eq_mask_loss2_}
\begin{split}
\mathcal{L}_{C} &= \mathbb{E}_{z,x\sim p(z,x)}[L(y\circ c+x\circ(1-c)) - \lambda \norm{c}]  \\
&= \mathbb{E}_{z,x\sim p(z,x)}[\sum_{i\in\mathcal{C}}l(y_i)+\sum_{i\in(\mathcal{H}-\mathcal{C})}l(x_i) - \lambda |\mathcal{C}|]\\
&= \mathbb{E}_{z,x\sim p(z,x)}[\sum_{i\in\mathcal{C}}l(y_i)- \lambda |\mathcal{C}|];  
\end{split}
\end{equation}
for a single sample $y$, the loss is the summation of local loss over $\mathcal{C}$. Therefore, the minimum is achieved when $\mathcal{C}$ covers the set of pixels with the smallest local loss values. 
Intuitively, the first term in Equation~\ref{eq_mask_loss} encourages the confidence map to have high response where the loss $L(y)$ is small, as $\mathcal{L}_{C}$ is expected to be smaller by choosing low-loss area from the generator output $y\circ c$ and replacing high-loss are with ground-truth $x\circ(1-c)$; the second term penalizes a trivial solution of all-zero confidence maps by encouraging the confident region to cover as much of the missing region as possible. Fig.~\ref{fig:loss_conf} illustrates how to compute $\mathcal{L}_{C}$. 

\begin{figure}[t]
\begin{center}
  \includegraphics[width=.9\linewidth]{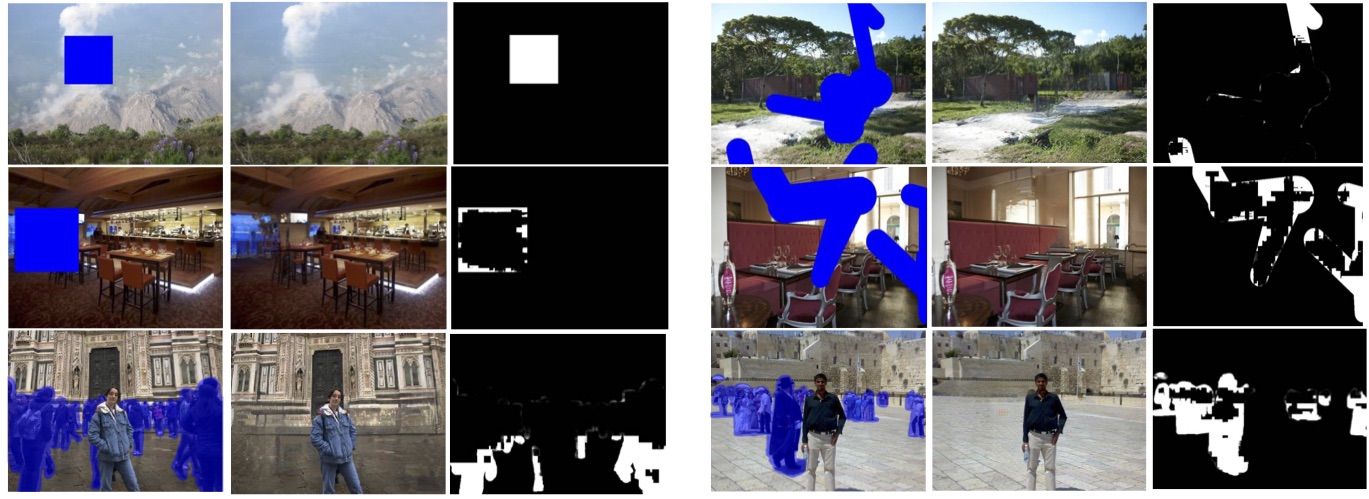}\\
\end{center}
  \caption{\small  Results and confidence maps. Blue masks indicate holes. Brighter pixels in confidence maps are of higher confidence. We use solid blue masks for synthetic samples and transparent ones for real object removal cases.}
\label{fig:confs}
\end{figure}
Fig.~\ref{fig:confs} shows examples of inpainting results and the corresponding confidence maps. We can see that the confidence maps tend to highlight good regions of the result. As one may expect, confident regions are often located close to the hole boundaries (e.g., the 2nd row). However, there are also other cases: 1) in easy cases like filling a hole in the sky, all generated pixels can have high confidence (e.g., 1st row, left); 2) flat regions tend to be more confident than highly-textured regions; 3) artifacts usually have low confidence (e.g., third row, right). 

\subsubsection{Iterative inpainting}
\label{sec_iterative_inpainting}
We can use the confidence decoder output to identify confident sub-regions in the inpainting result and run the inpainting model again and repeat the process. In each iteration, we set the confident pixels as "valid" pixels in the new input and set the remaining low-confidence regions as new holes for the next iteration to process. Overall, holes are shrinking as the iteration goes so that the network should be more certain about the generated result. 

\begin{figure}[t]
\begin{center}
   \includegraphics[width=.9\linewidth]{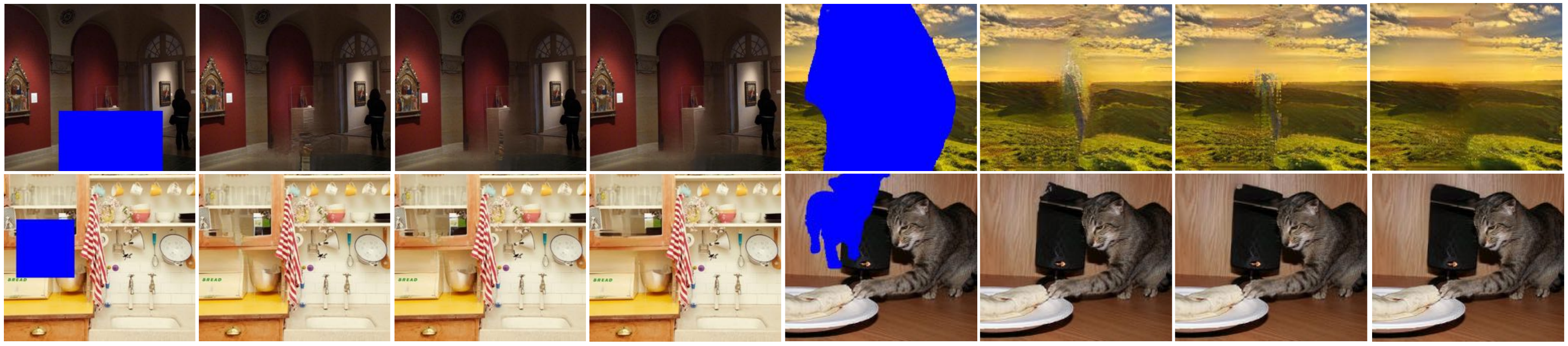}\\
\end{center}
   \caption{\small Inpainting results as iterations increase. 
   }
\label{fig:iterates}
\end{figure}

Algorithm~\ref{alg_iterative_test} describes the iterative inpainting process in details. In the first iteration, we initialize the completed image by filling the whole missing region with the generated content and set the pixels of which confidence is below 0.5 as the missing regions for the second iteration. From the second iteration, a pixel is replaced by new generated one if its confidence increases over the previous iteration. 
When training, we iterate twice. We also fix the number of iterations during the testing. There is no convergence issue because our algorithm always keep the current-best complete prediction inside the hole at every iteration. Fig.~\ref{fig:iterates} shows inpainting results in three consecutive iterations. As iteration goes, lines are connected and distorted area or artifacts are corrected.
\begin{algorithm}[t]
    \small
    \SetKwInOut{Input}{\textbf{Input}}
    \SetKwInOut{Output}{\textbf{Output}} 
    \Input{ 
    Incomplete image $z_1$,
    hole mask $m_1$, the number of iterations $T$ \\
    }
    \Output{ Completed image $y_T$\\
    }
    \BlankLine
    Set initial confidence map  $c_0=0.5m_1$\\
    \For {\ $t \in \left\{1,...,T\right\}$}{
        Get confidence map $c_t=C(z_t)\circ m_t$\\
        Get mask of regions to update $u_t=\mbox{Binarize}(c_t-c_{t-1}\circ m_t)$\\
        Update mask $m_{t+1} = m_t-u_t$\\
        \uIf { $t = 1$} {
            Initialize completed image $y_1 = z_1 + G(z_1)\circ m_t$\\
        }
        \Else{
        Update competed image $y_t = G(z_t)\circ u_t + y_{t-1}\circ (1-u_t)$\\
        }
        Update incomplete image $z_{t+1}=y_t\circ m_{t+1}$
    }
    \caption{\small $G(\cdot)$: generator; $C(\cdot)$: confidence decoder}\label{alg_iterative_test}
\end{algorithm}

\subsection{Guided upsampling}
\label{sec_guided}
Our iterative inpainting model is trained on low-resolution (LR) ($256 \times 256$) so it is not ideal to directly apply it to high-resolution (HR) inputs. To solve this issue, we propose a guided inpainting upsampling network to generate a HR inpainting result given a LR inpainting result. We propose a new architecture extending the contextual attention module~\cite{yu2018generative} which can match and use feature patches from valid surrounding areas to help synthesize the hole pixels.

As illustrated in Fig.~\ref{fig:overall_split1}~(b), the proposed guided upsampling network consists of two shallow networks, one for learning patch similarity and the other for image reconstruction. Their feature maps are of different sizes, but we can split them into an equal number of patches using different patch sizes so that patches of the similarity network feature map have 1:1 correspondence to patches of the reconstruction network feature map which allows us to use shared indices to represent patches. Let $\mathcal{H}$ and $\mathcal{V}$ to be the index set of patches inside the holes and the valid patches, respectively. Valid patches are those with at least one pixel outside the holes, and others are taken as patches inside the hole. The patch similarity network calculates the cosine similarity $s_{ij}$ between a pair of patch $i, j$.  
The reconstruction network is a shallow encoder-decoder network with skip connections from each layer of the encoder to the mirrored layer of the decoder. 
Before converting an HR feature map (of the HR input size) into an HR inpainting result, each feature patch inside the holes is replaced with a weighted sum of valid patches. Let $\phi_i$ to be an HR feature patch. The patch replacement in the HR feature maps can be summarized as follow:
\begin{equation}
\phi_i = \sum_{j\in \mathcal{V}} s^{'}_{ij} \phi_j, \ \ i \in \mathcal{H},
\end{equation}
where $s^{'}_{ij}$ is the softmax of $s_{ij}$: 
\begin{equation}
s^{'}_{ij} = \frac{\exp(s_{ij})}{\sum_{j\in \mathcal{V}} \exp(s_{ij})}.
\end{equation}
Then the HR feature maps are transformed to an output image by two convolution layers (``ToRGB" in Fig.~\ref{fig:overall_split1}~(b)). The loss on the HR output is a combined L1 and adversarial loss, the same as in Equation~\ref{eq_image_loss}. As mentioned earlier, we take the patches with at least one valid pixel as valid patches. For missing regions reconstructed by these partially valid patches, we simply take them as holes and run the previously described iterative inpainting model one more time. 
\begin{figure}[t]
\begin{center}
  \includegraphics[width=.9\linewidth]{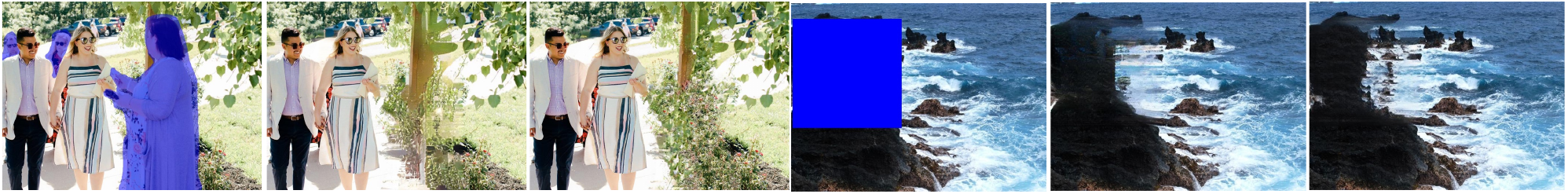}\\
  \makebox[\linewidth][s]{\scriptsize{{ } Input { } (a) { } (b) { } Input { } (a) { } (b) { }}}
\end{center}
  \caption{\small Effect of guided upsampling network. (a) results obtained by the iterative inpainting model on original size; (b) running the iterative model on half size and using guided upsampling network to upsample to the original size. }
\label{fig:upsample_example}
\end{figure}
By separating high-level similarity learning and low-level texture reconstruction, the proposed guided upsampling network can generate inpainting results that are both semantically reasonable and visually realistic, as shown in Fig.~\ref{fig:upsample_example}. 

\begin{figure}[t]
\begin{center}
  \includegraphics[width=.8\linewidth]{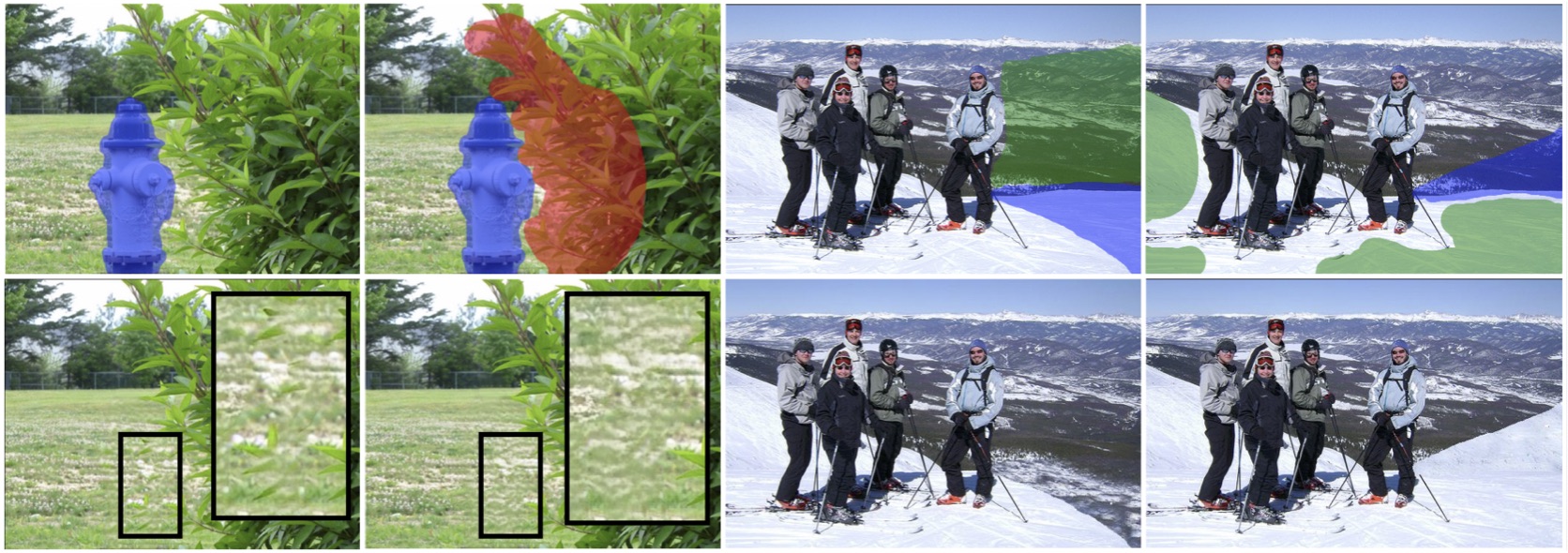}
\end{center}
  \caption{\small Effect of user control. The blue masks indicate holes, red and green ones respectively indicate regions avoid or specified to be used for filling the holes. The second example shows that our method with user control can also be applied to interactive image layout manipulation. 
  }
\label{fig:control}
\end{figure}

Another advantage of guided upsampling network is that it allows some user control over the results. As the contents inside the holes are constructed only using patches in $\mathcal{V}$, we can adjust the results by removing from or adding to $\mathcal{V}$. For example, as shown in Fig.~\ref{fig:control}, we can exclude the region that we do not want to copy, or specify a reference region used for filling the hole. 

\section{Experiments}
\subsection{Implementation details}
We implement our method using Python and Pytorch. Detailed network architectures can be found in the supplementary material\footnote{https://zengxianyu.github.io/iic}. We train the models with Adam~\cite{kingma2014adam} optimizer; the learning rate set to 0.0001. The training batch size is 64. As the Places2 is much larger than the saliency dataset, we sample an equal number of images from Places2 dataset and saliency dataset to constitute each batch to prevent the model from ignoring the scarce samples. 
We use $256\times 256$ patches for training the iterative inpainting model and train the guided upsampling network to upsample $256\times 256$ results to $512\times 512$. 
We randomly take 400 images from the training split of Places2 and the saliency dataset as validation samples and generate holes on them as described in Sec.~\ref{sec_data}. The model is trained until the PSNR on this validation set does not increase. The 1,000 testing images are kept unseen during training. When testing, the number of iterations for iterative inpainting is set to 4. 

\subsection{Comparison with state-of-the-art methods}
We evaluate quantitative scores and visual quality of two variants of our method: \textit{i.e.} \textbf{Ours*}: the iterative inpainting model running on original input without guided upsampling 
and \textbf{Ours}: the iterative inpainting model running on $2\times$ downsampled input and then using the guided upsample model to obtain the results of original size. We compare our methods with four state-of-the-art methods: Global\&Local~\cite{iizuka2017globally}, PatchMatch~\cite{barnes2009patchmatch}, GConv~\cite{yu2019free} and EdgeConnect~\cite{nazeri2019edgeconnect}. Comparison with more methods can be found in supplementary material. 

\subsubsection{Quantitative evaluation}
We evaluate two variants of our method and state-of-the-art methods on the test set of 1,000 images with object shaped holes. These images are of various size, from short side 256 to long side 1024. For random 500 images of them, we exclude salient objects from holes to simulate the case of distracting objects behind the main objects. For the rest 500 images the holes are placed randomly. 
For fair comparisons with previous methods, we also evaluate on the standard Places2 validation set resized to 256$\times$256 with 128$\times$128 center square holes as in most previous methods \textit{e.g.}~\cite{iizuka2017globally,yu2018generative,zeng2019learning}, and irregular holes as in~\cite{yu2019free,xiong2019foreground}. 
We use L1 loss, PSNR, and SSIM as they are most commonly used metrics in image inpainting. Tab.~\ref{table_metric} shows quantitative comparisons of our method with state-of-the-art
methods. Both variants of our method compare favourably against previous methods. Without guided upsampling and running on original resolution, \textbf{Ours*} model tends to generate smoother results, which are favored by these scores at per-pixel basis. 
\begin{table}[t]
\setlength{\tabcolsep}{1.5pt}
\caption{\small Quantitative evaluation and user preference of various methods. P.c.: preference count in user study. }
\label{table_metric}
\small
\begin{center}
\begin{tabular}{ccc|ccc|ccc|ccc|c}
\hline
\multicolumn{3}{c|}{\multirow{2}{*}{Method}} & \multicolumn{3}{c|}{Object shaped holes} & \multicolumn{3}{c|}{Irregular holes (Places2)}& \multicolumn{3}{c|}{Square holes (Places2)}&User study\\
\multicolumn{3}{c|}{}& L1 Loss &PSNR &SSIM& L1 Loss &PSNR &SSIM& L1 Loss &PSNR &SSIM&P. c.\\
\hline
\multicolumn{3}{c|}{\cite{barnes2009patchmatch}} & .0273 & 25.64 & .8780 & .0288 & 22.87 & .8549 & .0432 & 19.19 & .7922 &13\\
\multicolumn{3}{c|}{\cite{iizuka2017globally}} & .0292& 24.23& .8653 & .0385 & 20.95 & .8185 & .0386 & 20.16 & .7950&2\\
\multicolumn{3}{c|}{\cite{yu2019free}} & .0243 & 26.07 & .8803  & .0245 & 24.31 & .8718 & .0430& 19.08& .7984 &8\\
\multicolumn{3}{c|}{\cite{nazeri2019edgeconnect}} & .0246 & 26.24 & .8871 & .0221 & 24.78 & .8701 & .0368 & 20.30 & .8017&10\\
\hline
\multicolumn{3}{c|}{Ours*}& .0194 & 28.20 & .8985  & .0203 & 25.43 & .8828& .0361& 20.21& .8130 &62\\
\multicolumn{3}{c|}{Ours} & .0205 & 27.67 & .8949  & 0220 & 24.70 & .8744 & .0384& 19.69& .8063 &80\\
\hline
\end{tabular}
\end{center}
\end{table}
\begin{figure}[t]
\begin{center}
   \includegraphics[width=.9\linewidth]{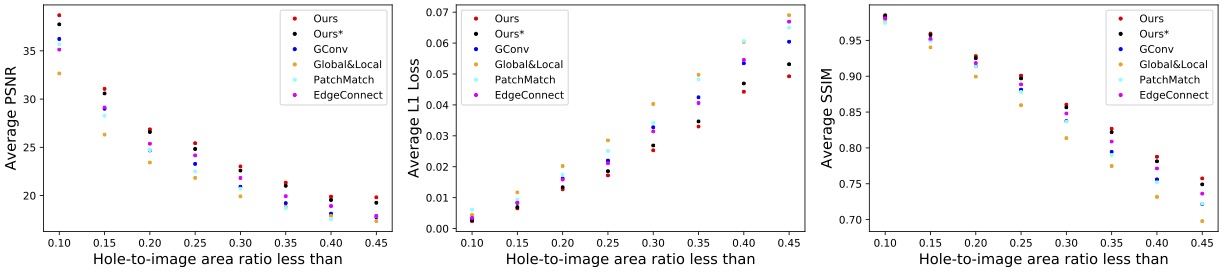}
\end{center}
   \caption{\small Quantitative comparisons when hole size varies. }
\label{fig:ablation_holesize}
\end{figure}
To validate superiority of our method on filling large holes, we show scores measured on images with holes of different sizes in Fig.~\ref{fig:ablation_holesize}. The X-axis represents the range of hole-to-image area ratio and Y-axis represents average L1 loss, SSIM and PSNR over all samples of which the hole-to-image area ratio are in the corresponding range. For example, the first column is averaged over samples whose hole-to-image ratio is less than $0.1$ and the second column is averaged over those greater than $0.1$ but less than $0.15$. For small holes, all methods perform almost equally well. It is increasingly more difficult to fill holes when their size grow. So the SSIM, PSNR of all methods  decrease and L1 loss increases as the hole-to-image ratio increases. When it comes to larger holes, our method performs better. 

\subsubsection{Visual quality}
\label{sec_visual}
\begin{figure*}[t]
\begin{center}
   \includegraphics[width=.9\linewidth]{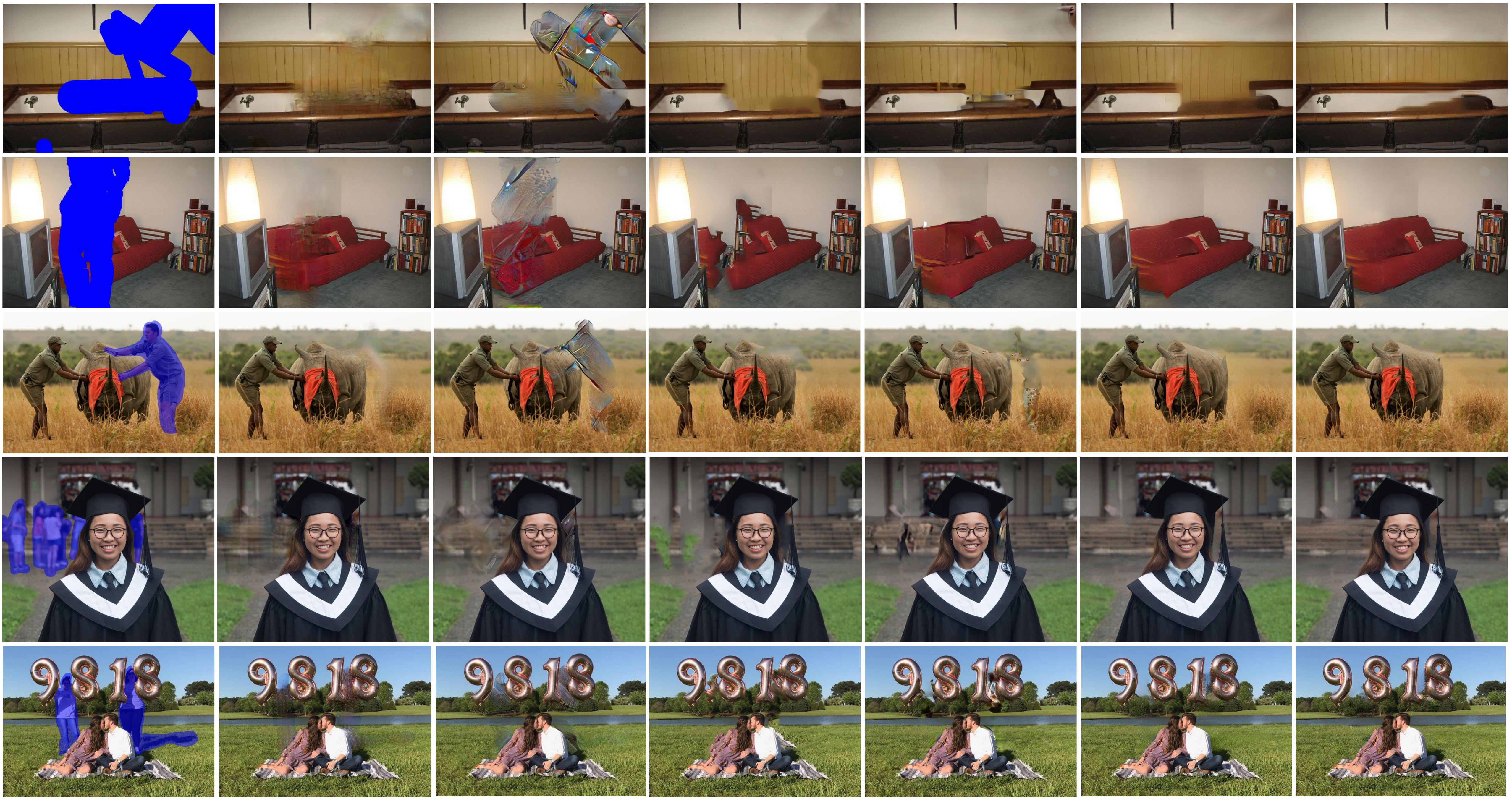}\\
   \makebox[\linewidth][s]{\scriptsize{ { } { }Input {}EdgeConnect {}Global\&Local{} PatchMatch{} GConv { } Ours* { } Ours{ } { } }}
\end{center}
   \caption{\small Visual comparison on mid-resolution images. Zoom-in to see the details. Images are compressed due to limited submission file size. More results can be found in the supplementary material. }
\label{fig:visual}
\end{figure*}
Fig.~\ref{fig:visual} shows visual comparisons of two variants of our method and state-of-the-art methods on synthetic samples and real object removal tasks~\cite{remove_coworker,remove_people,remove_people_balloon}. 
As shown in the figure, existing deep learning based methods do not work well in real requests. They often generate artifacts removing a large object. PatchMatch can generate clear texture, however, since it does not have a semantic understanding of input images, its results are not semantically reasonable. Both \textbf{Ours*} and \textbf{Ours} can provide reasonable alternatives for the region to remove. \textbf{Ours*} tends to generate smoother results. In comparison, by reconstructing LR results using patches from HR inputs, \textbf{Ours} method is better at keeping fine-grained details. Its results are similar to PatchMatch in terms of texture but more reasonable in terms of structure. 

To evaluate visual quality of our method, we conduct a user study on 25 real object removal cases collected from object removal requests on the Web. 
All images are resized to make the short side equal to 512. Each input image with a marked region to remove and the results of different methods are shown in random order to 11 users 
and we ask them to select a single best result. Each combination of input and results are shown twice, and a valid vote is counted only when a user selects the same result twice. Finally we collect 175 valid votes. The user study results are shown in Tab.~\ref{table_metric}. Both variants of our method are preferred more than previous methods. \textbf{Ours} model with guided upsampling tends to generate results that are less smooth and with more clear texture, which are often favored by users. 

\subsection{Ablation study}
\begin{table}[t]
\setlength{\tabcolsep}{2pt}
\caption{\small Effect of each component. IT: iterative inpainting; CF: confidence feedback; RT: realistic training data as described in~\ref{sec_data}; GU: guided upsampling. \textit{PC} represents preference counts in user study. Time is measured in seconds on $512\times512$ input. \textit{User ctrl.} indicates whether a method allows user control.  }
\label{table_ablation_score}
\small
\begin{center}
\begin{tabular}{cccc||ccc|c|c|c}
\hline
IT & CF & RT & GU & L1 Loss &PSNR &SSIM &PC&Time&User ctrl.\\
\hline
& & & & .0205 & 27.79 & .8903&-&.064&$\times$\\
$\surd$& & & & .0204 & 27.57 & .8925&-&.301&$\times$\\
$\surd$& $\surd$ & & & .0200 & 28.06 & .8952&-&.323&$\times$\\
$\surd$& $\surd$ & $\surd$& & .0194 & 28.20 & .8985&62&.323&$\times$\\
$\surd$& $\surd$ & $\surd$& $\surd$& .0205 & 27.67 & .8949 &80&.182&$\surd$\\
\hline
\hline
\multicolumn{4}{c||}{Confidence$>0.5$}&.0033& 36.38& .981 &- &- &-\\
\multicolumn{4}{c||}{Confidence$\le0.5$}&.0165& 30.00& .923 &- &- &-\\
\hline
\end{tabular}
\end{center}
\end{table}
First, to validate the proposed confidence prediction mechanism, we separately evaluate the results of high-confidence ($>0.5$) and low-confidence ($\le0.5$) regions inside the hole. The results are in the bottom two rows of Tab.~\ref{table_ablation_score}, which indicates that the prediction in high-confidence regions are significantly better than low-confidence regions. 
We show the effect of realistic training data, iterative inpainting, and guided upsample model in the first to the third rows of  Tab.~\ref{table_ablation_score}. The first row corresponds to our baseline model without confidence decoder trained on the Places2 dataset using irregular and square holes. 
The second row shows the effect of conducting progressive inpainting (IT) in a predefined boundary-to-center manner. For this setting, we evenly split each hole into four parts based on distance transformation and run the baseline model four times for each input. Each time we fill the part closest to the hole boundaries and update the hole mask accordingly. The third row shows the iterative inpainting method with confidence feedback (CF) trained on Places2 using irregular and square holes. By predicting confidence map, it can automatically correct wrong inpainted pixels and gradually improve the results, which yields better performance than predefined progressive inpainting in terms of quantitative scores. The sixth row corresponds to \textbf{Ours*} model described in previous sections. The comparison between the third and the fourth row shows the effect of including realistic training (RT) samples. 
\begin{figure}[t]
\begin{center}
   \includegraphics[width=.8\linewidth]{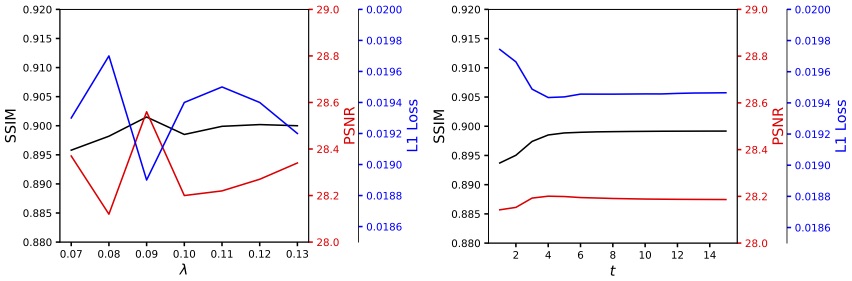}
   \makebox[\linewidth][s]{\scriptsize{{ } (a) { } (b) { }}}
\end{center}
   \caption{\small (a) Sensitivity analysis of $\lambda$. (b) Effect of increasing test iterations. }
\label{fig:ablation_test_iters}
\end{figure}
\begin{figure}[t]
\begin{center}
  \includegraphics[width=.9\linewidth]{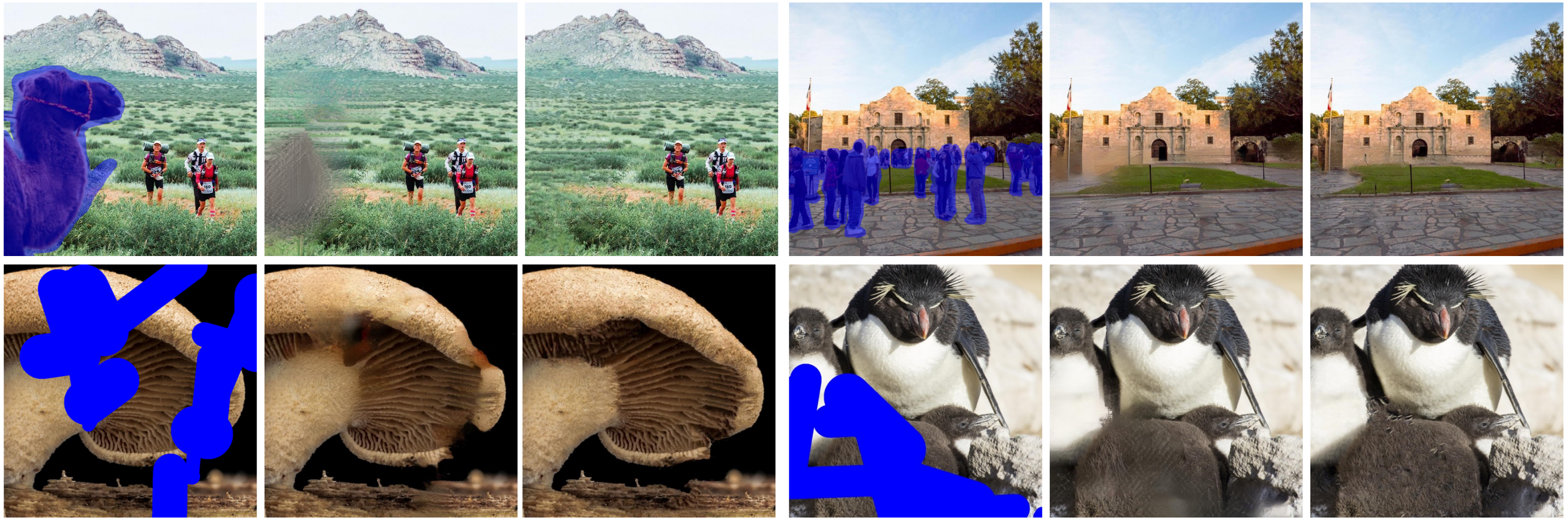}\\
  \makebox[\linewidth][s]{\scriptsize{{ }{ } Input { } Ours* { } Ours { } Input { } Ours* { } Ours { }{ }}}
\end{center}
   \caption{\small Inpainting results on $1024\times1024$ images. Zoom-in to see the details. Images are compressed due to limited submission file size. 
   }
\label{fig:visual1024}
\end{figure}
All the variants discussed above output results of the same size as the input. So when evaluating these models, we give them the original input and do not apply post processing on their results. To analyze the effect of guide upsampling, we first run the proposed iterative inpainting method on $2\times$ downsampled input images, and then upsample the results to the original resolution using guided upsampling. This corresponds to \textbf{Ours} described previously, and the effect is shown in the fifth row of Tab.~\ref{table_ablation_score}. By running the iterative process on the downsampled input, it significantly cuts down the overall run-time. 

The sensitivity analysis of $\lambda$ (Eqn.~\ref{eq_mask_loss}) is shown in Fig.~\ref{fig:ablation_test_iters}~(a). The performance is not very sensitive to $\lambda$ when it changes in a small range. For example, for $\lambda \in [0.7, 0.13]$, PSNR is in $[28.1, 28.5]$. 
Fig.~\ref{fig:ablation_test_iters}~(b) shows the effect of increasing test iterations. More test iterations generally lead to better scores, especially in the first four iterations. We fix the number of iterations to $4$ during testing.

Fig.~\ref{fig:visual1024} shows inpainting results of \textbf{Ours*} and \textbf{Ours} on input images of size $1024\times1024$. As the guided upsampling network lifts the LR result to HR by utilizing features from a HR input, it brings the details from existing contents to generated contents, resulting in a more visually pleasant HR output. It also can be reflected from the user study, in which \textbf{Ours} is preferred by users more frequently. 
However, reconstructing with existing patches is a constraint on generation, making it less free and difficult to restore exactly the original content in the missing region. As a result, \textbf{Ours} has lower quantitative scores than \textbf{Ours*} as shown in the last row of Tab.~\ref{table_ablation_score}. 

\section{Conclusion}
We propose a high-resolution image inpainting method for large object removal. 
Our model predicts the inpainting result as well as its confidence map, which is used to revise unsatisfactory regions in an iterative manner. To improve visual quality for high-res inputs, we first obtain a low-res result and then reconstruct it using high-res neural patches. 
Furthermore, we collect a large object masks dataset and synthesize realistic training samples that simulate realistic user inputs. Experiments show that our method outperforms existing methods and achieves better visual quality. 

\section*{Acknowledgements}
The paper is supported in part by National Key R\&D Program of China under Grant No. 2018AAA0102001, National Natural Science Foundation of China under grant No. 61725202, U1903215, 61829102, 91538201, 61771088,61751212, Fundamental Research Funds for the Central Universities under Grant No. DUT19GJ201, Dalian Innovation leader’s support Plan under Grant No. 2018RD07.

%
%
\bibliographystyle{splncs04}
\bibliography{egbib}
\end{document}